\documentclass[journal]{IEEEtran}

\ifCLASSINFOpdf
\else
   \usepackage[dvips]{graphicx}
\fi
\usepackage{url}

\usepackage{graphicx}
\usepackage{amsmath}
\usepackage{siunitx}
\usepackage{amssymb}
\usepackage{pgfplots}
\usepackage{cite}
\usepackage[pagebackref=true,breaklinks=true,letterpaper=true,colorlinks,citecolor=blue,linkcolor=blue,bookmarks=false]{hyperref}

\newcommand{\ve}[1]{\mathbf{#1}} %

\newcommand{\ra}[1]{\renewcommand{\arraystretch}{#1}}
\newcolumntype{x}[1]{>{\centering\arraybackslash}p{#1pt}}

\newlength\savewidth\newcommand\shline{\noalign{\global\savewidth\arrayrulewidth
		\global\arrayrulewidth 1pt}\hline\noalign{\global\arrayrulewidth\savewidth}}

\def\ie{i.e.,~}
\def\eg{e.g.,~}

\def\vs{vs.~}

\definecolor{mangotango}{rgb}{1.0, 0.51, 0.26}

\begin{document}

\title{CondNet: Conditional Classifier for Scene Segmentation}

\author{Changqian~Yu,
Yuanjie~Shao,
Changxin~Gao,
Nong~Sang
\thanks{Manuscript received January 18, 2021; revised March 10, 2021; accepted March 13, 2021. Date of publication April 1, 2021; date of current version April 28, 2021. This work is supported by the National Natural Science Foundation of China under Grants 61433007, 61876210 and 61901184. The associate editor coordinating the review of this manuscript and approving it for publication was Prof. Yue Deng. \textit{(Corresponding author: N. Sang)}}
\thanks{The authors are with the Key Laboratory of Image Processing and Intelligent Control, School of Artificial Intelligence and Automation, Huazhong University of Science and Technology, Wuhan 430074, China (e-mail: changqian\_yu@hust.edu.cn; shaoyuanjie@hust.edu.cn; cgao@hust.edu.cn; nsang@hust.edu.cn).}
\thanks{Digital Object Identifier 10.1109/LSP.2021.3070472}
}

\markboth{IEEE SIGNAL PROCESSING LETTERS, VOL. 28, 2021}
{YU \MakeLowercase{\textit{et al.}}: CONDNET: CONDITIONAL CLASSIFIER FOR SCENE SEGMENTATION}
\maketitle

\begin{abstract}
The fully convolutional network (FCN) has achieved tremendous success 
in dense visual recognition tasks, 
such as scene segmentation.
The last layer of FCN is typically 
a global classifier ($1\times1$ convolution) 
to recognize each pixel to a semantic label.
We empirically show that
this global classifier, 
ignoring the intra-class distinction, 
may lead to sub-optimal results.

In this work, we present a conditional classifier
to replace the traditional global classifier, 
where the kernels of the classifier 
are generated dynamically 
conditioned on the input.
The main advantages of the new classifier consist of:
(i) it attends on the intra-class distinction,
leading to stronger dense recognition capability;
(ii) the conditional classifier is simple and flexible 
to be integrated into almost arbitrary FCN architectures  
to improve the prediction.
Extensive experiments demonstrate 
that the proposed classifier 
performs favourably against the traditional classifier
on the FCN architecture.
The framework
equipped with the conditional classifier
(called CondNet) 
achieves new state-of-the-art performances on two datasets.
The code and models are available at \url{https://git.io/CondNet}.

\end{abstract}


\IEEEpeerreviewmaketitle

\section{Introduction}
\label{sec:intro}

\IEEEPARstart{S}{cene} segmentation~\cite{Huang-SPL-Soft-IoU-2019, Li-SPL-Proposal-2019, Arnab-SPM-CRF-2018, Zhou-WWW-Multi-scale-2019, Wang-CVPR-DSR-2020} 
is a fundamental and challenging task in visual recognition, 
aiming to recognize each pixel 
into a semantic category, 
providing comprehensive scene understanding.
It has extensive downstream applications, 
\eg autonomous driving~\cite{Badrinarayanan-PAMI-SegNet-2017, Kitti, Cityscapes, Zhou-ASC-AGLNet-2020}, 
human-machine interaction~\cite{Liang-PAMI-lip-2018}, 
and augmented reality.

In recent years,
with the development of deep neural networks~\cite{Krizhevsky-NIPS-Imagenet, Simonyan-ICLR-VGG-2015, He-CVPR-ResNet-2016, Huang-CVPR-DenseNet-2017},
the fully convolutional network (FCN)~\cite{Long-CVPR-FCN-2015}
has achieved tremendous success and been the dominant solution
in the scene segmentation task.
In the original FCNs, 
the whole architecture 
composes of a \emph{feature extractor} and a \emph{classifier} 
(the last layer for prediction).

Feature extractors
have been widely studied for 
embedding powerful feature representation.
Various designs have been proposed to 
extract effective discriminative features:
(i) larger receptive field, \eg dilation/deformable convolutions~\cite{Chen-ICLR-Deeplab-2016, Dai-ICCV-DCN-2017, Yu-ECCV-RepGraph-2020},
(ii) multi-scale representation via pyramid methods, \eg PPM\cite{Zhao-CVPR-PSPNet-2017}, ASPP\cite{Chen-Arxiv-Deeplabv3-2017} and MDCCNet\cite{Zhou-WWW-Multi-scale-2019},
(iii) adaptive aggregation via attention mechanism, \eg self-attention~\cite{Fu-TNNLS-DANet-2020, Wang-CVPR-Nonlocal-2018, Yu-CVPR-CPNet-2020} and channel attention mechanism~\cite{Yu-CVPR-DFN-2018, Hu-CVPR-SEnet-2017,Yu-ECCV-BiSeNet-2018,Yu-ARXIV-BiSeNetV2-2020,Zhou-ASC-AGLNet-2020}.

\begin{figure}[t]
\footnotesize
\centering
\renewcommand{\tabcolsep}{1pt} %
\ra{1} %
\begin{center}
\begin{tabular}{cc}
\includegraphics[width=0.5\linewidth]{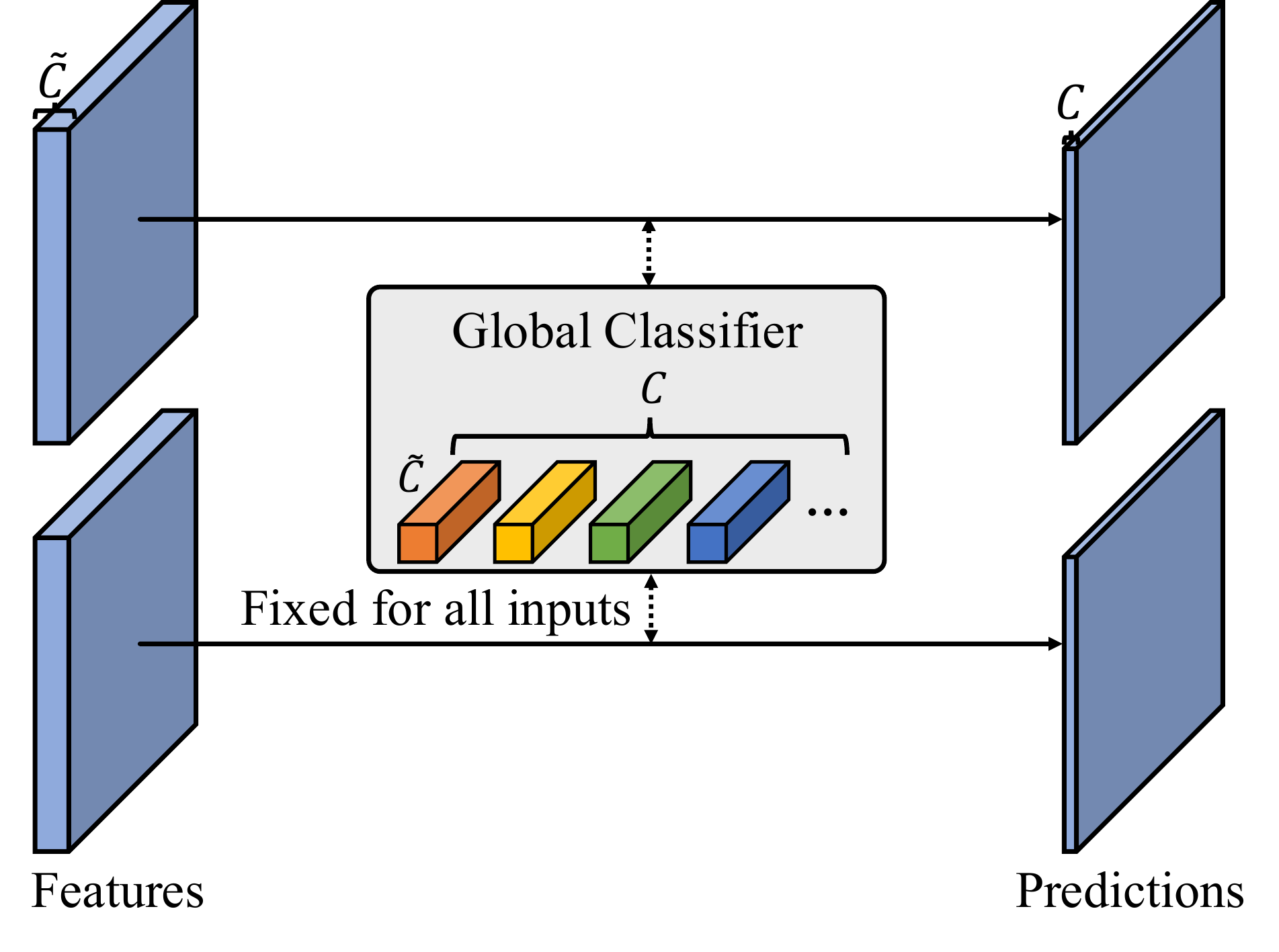} &
\includegraphics[width=0.5\linewidth]{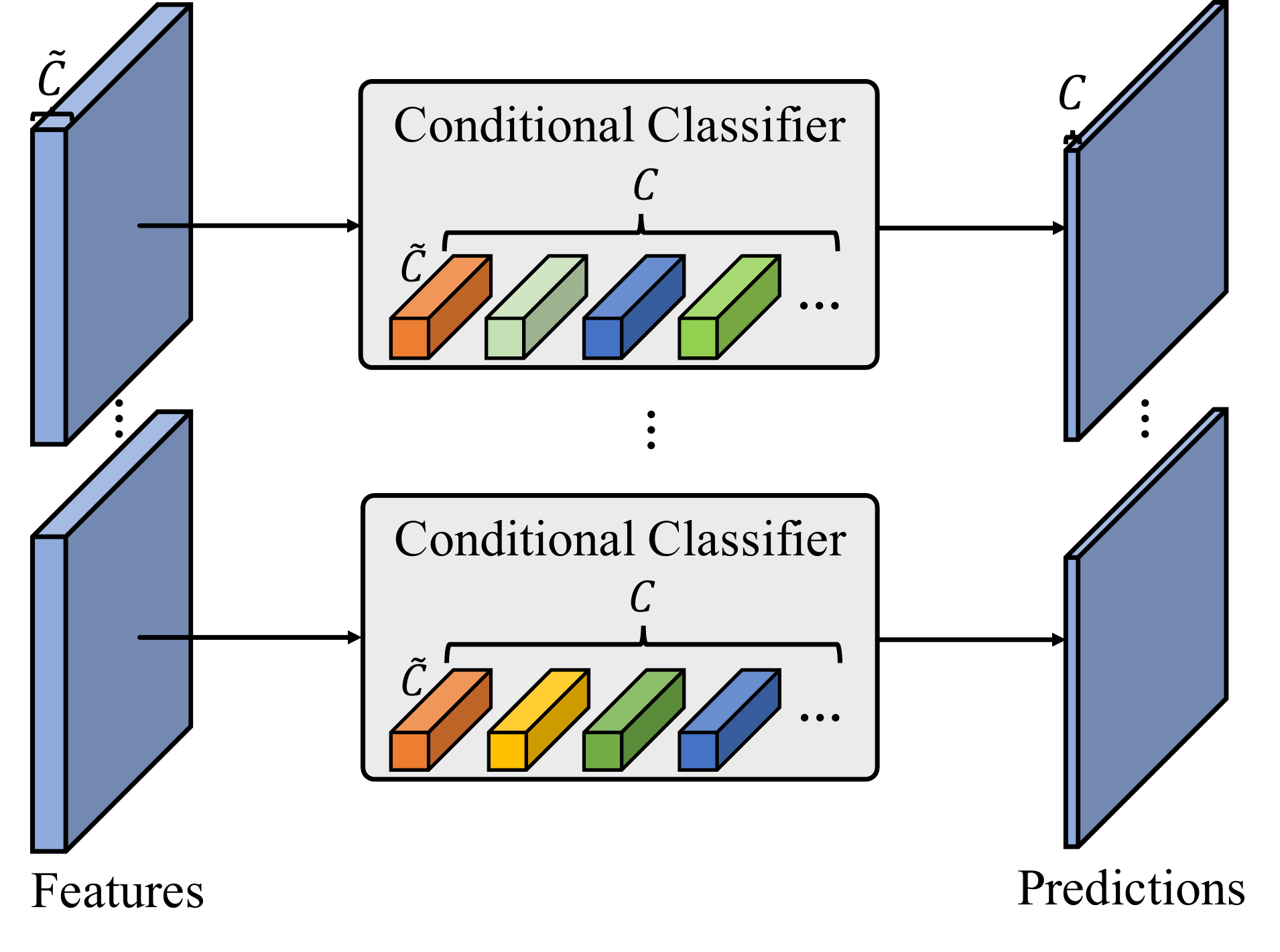} \\
(a) Global Classifier &
(b) Conditional Classifier \\
\end{tabular}
\end{center}
\caption{\textbf{Illustration of global classifier and conditional classifier.}
Global classifier stays fixed for all inputs.
Conditional classifier generates diverse kernels for different samples conditioned on the inputs.}
\label{fig:fig1}
\end{figure}

In contrast, 
the classifier has been studied in relatively few works.
The traditional classifier 
performs the kernel correlation 
on each position of the feature map 
to obtain the desired pixel-wise prediction.
In the training phase,
the kernel is learned thorough the whole training samples.
In the evaluation phase,
the learned kernels stay fixed 
and are applied to the feature maps
to predict the semantic maps.
We call this type of classifier as the global classifier,
which attempts to seek a \textbf{global class center} 
to recognize all the variation of different samples,
as shown in Figure~\ref{fig:fig1} (a).

The drawback of the global classifier is 
its limited capability 
to handle the intra-class distinction.
In some complex scene,
the diverse samples of the same semantic category
may have very different appearances 
(which we call intra-class distinction/variation).
It is common that the traditional classifier
is easy to mis-recognize these pixels 
of the same category but different appearances into
different categories 
since it is global for the majority of pixels of one category.

In this work, 
we present a conditional classifier for 
pixel-wise recognition, 
replacing the global classifier 
used extensively in previous works.
The main idea behind our work
is to generate the sample-specific kernels
with parameters adapted to 
the particular patterns within an input sample, 
which can handle the intra-class distinction,
as shown in Figure~\ref{fig:fig1} (b).

Our conditional classifier consists of two parts:
the \emph{class-feature aggregation module}
and
the \emph{kernel generation module}.
The class-feature aggregation module
tends to aggregate the features of each semantic category
via weighted average.
It is expected that 
the weighted average manner 
can capture the distinction of the same category within one sample
as the \textbf{sample-specific class center} (instead of the global class center).
The kernel generation module
dynamically generates the sample-specific kernels 
conditioned on the sample-specific class center.
The generated sample-specific kernels 
are applied to the input sample
to predict the semantic masks. 

There are several merits of the proposed conditional classifier: 
(i) The conditional classifier attends on the sample-specific distinction of each category to learn a more discriminative classifier.
(ii) The conditional classifier can be seamlessly
incorporated into almost arbitrary FCN architectures, 
replacing the global classifier ($1\times1$ convolution).

Extensive evaluations on the scene segmentation task
demonstrate that 
the conditional classifier performs favourably against 
the traditional classifier. 
The framework equipped with 
the conditional classifier (called \textbf{CondNet})
achieves new state-of-the-art performances 
on two challenging datasets.

\begin{figure}[t]
\footnotesize
\centering
\renewcommand{\tabcolsep}{1pt} %
\ra{1} %
\begin{center}
\begin{tabular}{c}
\includegraphics[width=0.95\linewidth]{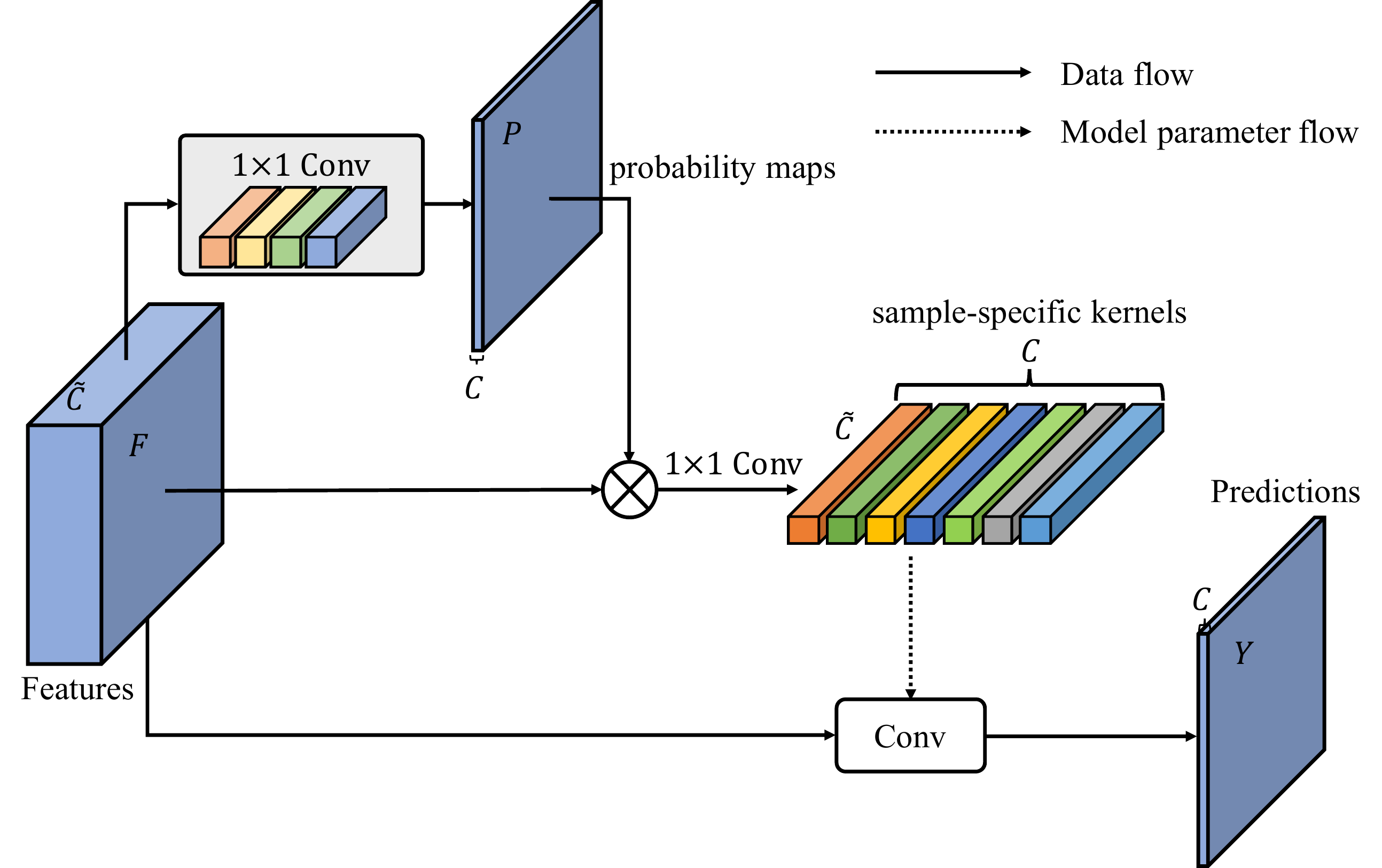} \\
\end{tabular}
\end{center}
\caption{\textbf{Structure of the proposed conditional classifier.}
The conditional classifier aggregates the features of each semantic category
as the sample-specific class centers 
guided by the coarse probability maps.
Then, the class centers are transformed to generate the sample-specific kernels
to recognize the features to the final predictions.
The probability maps are generated by a $1\times1$ convolution.}
\label{fig:structure}
\end{figure}

\section{Proposed Method}
\label{sec:method}
In this section,
we first revisit the global classifier used extensively in previous works.
Next,
we formulate the proposed conditional classifier.
Finally,
we describe the overall architecture with our proposed conditional classifier
and the corresponding loss functions.

\subsection{Revisiting Global Classifier}
A typical global classifier is 
a $1 \times 1$ convolution used as 
the last layer of the segmentation architecture,
as shown in Figure~\ref{fig:fig1} (a).
Consider an input feature $\mathsf{F} \in \mathcal{R}^{\tilde{C} \times H \times W}$ 
(the final output of the feature extractor), 
and the desired prediction $\mathsf{Y} \in \mathcal{R}^{C \times H \times W}$,
where $H$, $W$, $\tilde{C}$, $C$ denote the height, width, channel dimensions, and the number of semantic categories, respectively.

The global classifier performs a matrix-vector
multiplication on each position of $\mathsf{F}$:
\begin{align}
	\mathsf{Y} = \mathcal{H}(\mathsf{F}, \mathbf{W}),
\end{align}
where $\mathcal{H}$ is the global classifier ($1 \times 1$ convolution) with the kernels $\mathbf{W} \in \mathcal{R}^{C \times \tilde{C}}$, $\mathsf{Y}$ is the prediction maps, and $\otimes$ indicates the convolutional operation.
For each semantic category,
the classifier has a kernel $\ve{w}^{s} \in \mathcal{R}^{1 \times \tilde{C}}$, where $s \in \{1, 2, \dots, C\}$.
After training, the kernel stays fixed and 
is applied to recognize all different samples 
(that is why we call it global classifier).
Therefore,
the learned kernel is required to 
capture all the variation between different samples of the same category 
to output the correct prediction.
However, 
the different samples of the same category 
may have far different appearances, 
especially in some complex scene.
We argue that 
the global classifier is hard to 
capture all variation, 
thus leading to sub-optimal results.

In fact, 
for an input image,
the pixels of the same category
have more similar patterns 
due to belonging to the same scene.
In other words,
the pixels of the same category
in the same scene
have a \textbf{sample-specific class center},
which, intuitively, is easier to recognize these pixels 
than the global class center. 
This motivates us 
to propose the conditional classifier,
which generates the sample-specific kernels 
conditioned on the sample-specific class centers.

\subsection{Conditional Classifier}
The overall structure of the conditional classifier
is shown in Figure~\ref{fig:structure}.
It mainly consists of two parts:
\emph{class-feature aggregation module}
and 
\emph{kernel generation module}.

\paragraph{Class-feature aggregation}
The goal of the class-feature aggregation module
is to embed the sample-specific class center.
For one input sample,
it requires to aggregate 
all the features of the same category 
as the class center embedding.
The embedding $\mathsf{E}^s \in \mathcal{R}^{\tilde{C}}$ 
is defined as the weighted average of the features 
belonging to the category $s$, as formulated as follows:
\begin{align}
    \mathsf{E}^s = \frac{\sum_{j=0}^{N}{p}^s_j \mathsf{F}_j}{N},
\end{align}
where $N=H \times W$, $p^s \in \mathcal{R}^{H \times W}$ denotes the probability map belonging to category $s$.
Here, we use the coarse prediction masks of the segmentation network as the probability maps $\mathsf{P} \in \mathcal{R}^{C \times H \times W}$,
which are generated by a $1\times1$ convolution.

The sample-specific class center embeddings 
capture the variation of different pixels within one input sample.
It is easier to handle the particular patterns of one certain sample
than the global class center.

\paragraph{Kernel generation}
We use the projection $\mathcal{H}_{\theta}$ with kernels $\mathbf{W}_{\theta}$ to transform
the sample-specific class center embeddings
to the sample-specific kernels as:
\begin{align}
	\mathbf{W}_{\phi} = \mathcal{H}_{\theta}(\mathsf{E}, \mathbf{W}_{\theta}),
\end{align}
where $\mathbf{W}_{\phi}$ is the sample-specific kernels 
conditioned on the corresponding class center embeddings.
We use group $1 \times 1$ convolution as the projection.

The generated filters are finally applied to each position of the input features $\mathsf{F}$
as:
\begin{align}
  \mathsf{Y} = \mathcal{H}_{\phi}(\mathsf{F}, \mathbf{W}_{\phi})
  = \mathcal{H}_{\phi}(\mathsf{F}, \mathcal{H}_{\theta}(\frac{\mathsf{P} \times \mathsf{F}^{\top}}{N}, \mathbf{W}_{\theta})),
\end{align}
where 
$\mathcal{H}_{(\cdot)}$ indicates the convolution, while $\times$ is the matrix multiplication,
$N=H \times W$, $H$ and $W$ is the height and width of the $\mathsf{F}$.
In other words, 
each generated weight/kernel 
correlates on the feature maps to highlight 
the pixels belong to the same category 
as the prediction map.

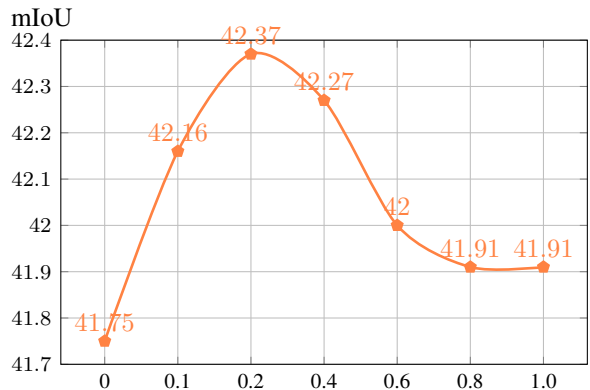
\begin{figure}[t]
\centering
\begin{tikzpicture}[baseline]
\begin{axis}[
	footnotesize,
	scale only axis,
	symbolic x coords={
	0, 0.1, 0.2, 0.4, 0.6, 0.8, 1.0},
	x post scale=1.4,
	xtick=data,
    nodes near coords, 
	nodes near coords align={vertical},
	every axis y label/.style={
at={(ticklabel cs:1.07)},rotate=0,anchor=near ticklabel, xshift=3em
},
	ymin=41.7, ymax=42.4,
	ylabel={mIoU},
	grid=major,
]
	\addplot[mangotango, mark=pentagon*, mark size=2pt, line width=1pt, smooth]
	coordinates{
	(0, 41.75)
	(0.1, 42.16)
	(0.2, 42.37)
	(0.4, 42.27)
	(0.6, 42.0)
	(0.8, 41.91)
	(1.0, 41.91)
	};
\end{axis}
\end{tikzpicture}
\caption{\textbf{Influence of the loss weights for the Soft Dice Loss.}
Different loss weights achieve similar results, indicating that the Soft Dice Loss is robust.
Here, we choose $\lambda = 0.2$ as default.}
\label{fig:loss_weight}
\end{figure}

\paragraph{Relation to other conditional architectures}
Different from normal networks,
conditional architectures can achieve 
dynamic kernels.
Dynamic filter networks~\cite{Xu-NIPS-DFN-2016} 
generates the convolution filters 
conditioned on the input.
PAC convolution~\cite{Su-CVPR-PAC-2019}
dynamically modifies the kernel with
an adapting kernel.
CondINS~\cite{Tian-ECCV-CondINS-2020} applys this design to the instance segmentation task,
generating the parameters of the mask sub-network
for each instance.
SVCNet\cite{Ding-CVPR-SVCNet-2019} 
learns a semantic correlation
dependent shape-variant context.
CondConv~\cite{Yang-NIPS-CondConv-2019} and Dynamic Convolution~\cite{Chen-CVPR-DynamicConv-2020}
learn a series of weights 
to mix the corresponding convolution kernels 
for each sample, increasing the model capacity.
Different from these methods,
we employ the explicit supervision 
on the
conditional generation process.

\subsection{Overall Architecture}
The proposed conditional classifier is flexible
to be integrated into almost arbitrary FCN architectures.
The architecture can replace the original classifier 
(the last layer of the architecture)
with the proposed conditional classifier directly.
The feature extractor outputs the embedding features,
then the coarse probability maps are predicted.
With the probability maps,
we can aggregate class features as the sample-specific class center embeddings,
then generate the sample-specific kernels 
to output the final predictions.

The overall loss function of the framework equipped with 
the conditional classifier can be formulated as:
\begin{align}
	\mathsf{L}_{overall} = \lambda \mathsf{L}_{prob} + \mathsf{L}_{seg},
\end{align}
where $\mathsf{L}_{prob}$ and $\mathsf{L}_{seg}$ denote
the loss of the coarse probability maps
and the loss of the conditional classifier, respectively.
We set $\lambda$ to $0.2$ in this work to balance these two losses.
We will give further comparisons
to discuss the influence of $\lambda$
at the experimental section.
$\mathsf{L}_{seg}$ is the Cross Entropy Loss, 
while $\mathsf{L}_{prob}$ is the Soft Dice Loss~\cite{Milletari-3DV-VNet-2016}
due to its effectiveness and stability in training
for class imbalance issue.
The Soft Dice Loss is defined as:
\begin{align}
	\mathsf{L}_{Dice} = 1 - \frac{2\sum_i^N{p_i q_i}}{\sum_i^{N}p_i^2 + \sum_i^{N}q_i^2 + \epsilon},
\end{align}
where $p_i\in \mathcal{R}^{C}$ is the probability vector of the probability maps, 
while while $q_i\in \mathcal{R}^{C}$ is a one-hot encoding vector of the corresponding ground truth masks, $N = H \times W$, $\epsilon$ prevents division by zero.

\begin{table}[t]
\centering
\setlength{\tabcolsep}{6.0pt}
\small
\renewcommand{\arraystretch}{1.4}
\caption{\textbf{Different loss functions for coarse probability maps}}
\label{tab:loss_function}
\begin{tabular}{l|c|c}
\shline
model, R50 & mIoU(\%) & picAcc(\%)\\ 
\shline
FCN baseline & 35.94 & 77.39 \\
\hline
w/o supervision & 41.75 & 79.89 \\
w/ BCE Loss & 40.43 & 79.67 \\
w/ Soft Dice Loss & \textbf{42.37} & \textbf{79.99} \\ 
\shline	
\end{tabular}
\end{table}	

\begin{table}[t]
\centering
\setlength{\tabcolsep}{5.0pt}
\small
\renewcommand{\arraystretch}{1.4}
\caption{\textbf{Comparison of conditional classifier and global classifier}}
\label{tab:cond_classifier}
\begin{tabular}{l|c|c|c|c}
\shline
\multicolumn{1}{c|}{classifier} & FCN & PSPNet & DeeplabV3 & DeeplabV3+ \\ 
\shline
\emph{global} &  35.94 & 41.13 & 42.42 & 42.72 \\
\emph{conditional} & 42.37 & 42.42 & 43.71 & 43.76 \\
\hline
$\Delta$ & +6.43 & +1.29 & +1.29 & +1.04 \\
\shline	
\end{tabular}
\end{table}

\section{Experiments}
We evaluate our approach on two scene segmentation datasets, \ie ADE20K and PASCAL-Context.
We perform a comprehensive ablation
on ADE20K dataset, and
report the comparisons 
with other methods on ADE20K and PASCAL-Context datasets.

\subsection{Setting}
\paragraph{Datasets}
ADE20K~\cite{Zhou-ADE-2016} is a scene understanding dataset,
containing $20$K training images and $2$K validation images. 
It has up to $150$ category labels for challenging scenes.

PASCAL-Context~\cite{PASCAL-Context} provides comprehensive scene understanding 
for both stuff and thing.
It can be divided into $4,998$ images for training 
and $5,105$ images for testing. 
The most common $59$ categories are used for evaluation.

\paragraph{Training}
The network is trained on 8 NVIDIA V100 GPUs with mini-batch 16 per GPU.
We adopt SGD optimizer with $0.9$ momentum and the initial learning rate of $4e^{-3}$ for ADE20K, $1e^{-3}$ for PASCAL-Context.
The ``poly'' learning rate~\cite{Chen-ECCV-Deeplabv3p-2018, Yu-CVPR-DFN-2018, Yu-CVPR-CPNet-2020} strategy is employed for the training process, 
in which the learning rate is multiplied by $(1 - \frac{iter}{max\_iter})^{0.9}$.
We use the synchronized batch normalization to train our networks.
The total training iterations are $160$K and $80$K for ADE20K and PASCAL-Context datasets, respectively.

The input image size is cropped to $512\times512$ for ADE20K, and $480\times480$ for PASCAL-Context.
Each image will go through a series of data augmentations,
containing random flipping, random scale ($[0.5, 2.0]$) for both datasets.

\paragraph{Testing}
We adopt the sliding-window evaluation strategy.
For the final results,
following~\cite{Zhao-CVPR-PSPNet-2017, Chen-ECCV-Deeplabv3p-2018, Yu-CVPR-CPNet-2020},
we average the predictions of multiple scaled ($[0.5, 1.75]$) and flipped inputs 
to improve the performance further. 
In addition, we adopt the pixel accuracy (pixAcc) and mean IoU (mIoU) metrics for ADE20K, and mIoU for PASCAL-Context.

\subsection{Ablation Study}
We perform ablative evaluations on ADE20K validation set 
to demonstrate the effectiveness of our approach.
We adopt the pre-trained ResNet-50~\cite{He-CVPR-ResNet-2016} as the backbone 
and the training iterations are $80$K.

\paragraph{Loss function}
Table~\ref{tab:loss_function} compares different loss functions
for the coarse probability maps.
We adopt the FCN based on ResNet-50 as our baseline.
Without explicit supervision,
the proposed classifier has achieved
$5.81$ mIoU increase.
As shown, the Binary Cross Entropy (BCE) Loss
achieves slightly worse results 
than the conditional classifier without supervision.
It is because
the majority of the pixels belong to the background for some certain category.
The imbalance issue is hard for the BCE Loss.
The Soft Dice Loss is designed to 
mitigate this issue.
Therefore,
this classifier with the Soft Dice Loss 
achieves
$6.43$ point gain.
Figure~\ref{fig:loss_weight}
illustrates the influence of the loss weights $\lambda$.
We choose $\lambda=0.2$ as default.

\paragraph{Conditional classifier \vs global classifier}
We apply the conditional classifier and global classifier
to different approaches,
including FCN, PSPNet, DeeplabV3, DeeplabV3+,
respectively.
The comparison of the dense recognition capability 
can be quantitatively measured by the performance contrast.
Table~\ref{tab:cond_classifier}
shows that the proposed conditional classifier
achieves better performance than the global classifier.
Specifically,
the conditional classifier improves 
FCN by $6.43\%$ mIoU,
PSPNet by $1.29\%$ mIoU,
DeeplabV3 by $1.29\%$ mIoU,
and DeeplabV3+ by $1.04\%$ mIoU,
respectively.
On the other hand,
Table~\ref{tab:cond_classifier}
also indicates
that the conditional classifier
is simple and flexible to 
integrate into the existing FCN architectures.

\begin{table}[t]
\centering
\setlength{\tabcolsep}{2.0pt}
\small
\renewcommand{\arraystretch}{1.4}
\caption{\textbf{Evaluation on
	the ADE20K validation set}}
\label{tab:ade20k}
\begin{tabular}{l|c|c|c|c}
\shline
\multicolumn{1}{c|}{model} & reference & backbone & mIoU($\%$) & picAcc($\%$)\\ 
\shline
UperNet~\cite{Xiao-UperNet-ECCV-2018}   & ECCV2018 & ResNet-101 & 42.66 & 81.01 \\
PSPNet~\cite{Zhao-CVPR-PSPNet-2017}    & CVPR2017 & ResNet-269 & 44.94 & 81.69 \\
PSANet~\cite{Zhao-ECCV-PSANet-2018}    & ECCV2018 & ResNet-101 & 43.77 & 81.51 \\
EncNet~\cite{Zhang-CVPR-EncNet-2018}    & CVPR2018 & ResNet-101 & 44.65 & 81.69 \\
CFNet~\cite{Zhang-CVPR-CFNet-2019}     & CVPR2019 & ResNet-101 & 44.89 & $-$ \\
ANL~\cite{Zhu-ICCV-ANL-2019}       & ICCV2019 & ResNet-101 & 45.24 & $-$ \\
OCRNet~\cite{Yuan-ECCV-OCRNet-2019}    & ECCV2020 & ResNet-101 & 45.28 & $-$ \\
APCNet~\cite{He-CVPR-APCNet-2019}    & CVPR2019 & ResNet-101 & 45.38 & $-$ \\
RGNet~\cite{Yu-ECCV-RepGraph-2020}     & ECCV2020 & ResNet-101 & 45.80 & 81.76 \\
CPNet~\cite{Yu-CVPR-CPNet-2020}    & CVPR2020 & ResNet-101 & 46.27 & 81.38 \\
DeeplabV3+~\cite{Chen-ECCV-Deeplabv3p-2018} & ECCV2018 & ResNet-101 & 46.35 & 82.40 \\
\hline
CondNet	  & & ResNet-101  & \underline{47.38} & \underline{82.49} \\
CondNet	  & & ResNest-101 & \textbf{47.54} & \textbf{82.51} \\
\shline	
\end{tabular}
\end{table}

\subsection{Results}
We employ the proposed conditional classifier to 
DeeplabV3+~\cite{Chen-ECCV-Deeplabv3p-2018} (called \textbf{CondNet}) to 
compare with other methods 
on two datasets: ADE20K and PASCAL-Context.

\paragraph{ADE20K}
The results of our method 
and other state-of-the-art methods are reported in Table~\ref{tab:ade20k}.
Based on ResNet-101~\cite{He-CVPR-ResNet-2016}, 
CondNet achieves $47.38\%$ mIoU and $82.49\%$ picAcc, 
significantly outperforming 
DeeplabV3+ by $1.03$ points,
CPNet by  $1.11$ points, RGNet by  $1.58$ points.
With more powerful backbone networks, ResNest-101~\cite{Zhang-ARXIV-ResNest-2020}, our CondNet achieves $47.54\%$ mIoU.
Besides,
we train the CondNet on the \texttt{train+val} set 
and submit the results on test set.
The CondNet based on ResNet-101 achieves
a final score of $0.5742$,
while the CondNet based on ResNest-101 achieves
$0.5754$.

\paragraph{PASCAL-Context}
Table~\ref{tab:pascal_cxt} reports the comparison results
of our networks and other state-of-the-art methods.
With ResNet-101, our method achieves $56.0\%$ mIoU
and outperforms 
OCRNet by $1.2$ points, APCNet by $1.3$ points, CFNet by $2.0$ points.
With ResNest-101, CondNet improves the mIoU to $57.0\%$ further.

\begin{table}[t]
\centering
\setlength{\tabcolsep}{4.0pt}
\small
\renewcommand{\arraystretch}{1.4}
\caption{\textbf{Evaluation on
	the PASCAL-Context validation set}}
\label{tab:pascal_cxt}
\begin{tabular}{l|c|c|c}
\shline
\multicolumn{1}{c|}{model} & reference & backbone & mIoU($\%$) \\ 
\shline
PSPNet~\cite{Zhao-CVPR-PSPNet-2017}    & CVPR2017 & ResNet-101 & 47.8  \\
DeeplabV3+~\cite{Chen-ECCV-Deeplabv3p-2018} & ECCV2018 & ResNet-101 & 48.3 \\
CCL~\cite{Ding-CVPR-CCL-2018}       & CVPR2018 & ResNet-101 & 51.6  \\
EncNet~\cite{Zhang-CVPR-EncNet-2018}    & CVPR2018 & ResNet-101 & 51.7  \\
DANet~\cite{Fu-CVPR-DANet-2019}     & CVPR2019 & ResNet-101 & 52.6  \\
SVCNet~\cite{Ding-CVPR-SVCNet-2019} & CVPR2019 & ResNet-101 & 53.2 \\
ANL~\cite{Zhu-ICCV-ANL-2019}       & ICCV2019 & ResNet-101 & 52.8  \\
CPNet~\cite{Yu-CVPR-CPNet-2020}     & CVPR2020 & ResNet-101 & 53.9 \\
RGNet~\cite{Yu-ECCV-RepGraph-2020}     & ECCV2020 & ResNet-101 & 53.9 \\
CFNet~\cite{Zhang-CVPR-CFNet-2019}     & CVPR2019 & ResNet-101 & 54.0 \\
APCNet~\cite{He-CVPR-APCNet-2019}    & CVPR2019 & ResNet-101 & 54.7 \\
OCRNet~\cite{Yuan-ECCV-OCRNet-2019}    & ECCV2020 & ResNet-101 & 54.8 \\
\hline
CondNet & & ResNet-101  & \underline{56.0} \\
CondNet & & ResNest-101 & \textbf{57.0} \\
\shline	
\end{tabular}
\end{table}

\section{Conclusion}
In this letter, we propose a conditional classifier 
to replace the traditional global classifier ($1\times1$ convolution for prediction)
in the FCN architecture.
For each input sample, this novel classifier aggregates the features of each category as the sample-specific class centers, and dynamically generates the corresponding kernels.
The kernels attend on the intra-class distinction, leading to stronger recognition capability.
The conditional classifier is easy and flexible to be integrated into almost arbitrary FCN architectures to improve the prediction results.
Finally, the framework equipped this classifier (called CondNet) achieves 
new state-of-the-art results on two challenging datasets.

\clearpage

\ifCLASSOPTIONcaptionsoff
  \newpage
\fi
\newpage
\bibliographystyle{IEEEtran}
\bibliography{reference}

\begin{thebibliography}{10}
\providecommand{\url}[1]{#1}
\csname url@samestyle\endcsname
\providecommand{\newblock}{\relax}
\providecommand{\bibinfo}[2]{#2}
\providecommand{\BIBentrySTDinterwordspacing}{\spaceskip=0pt\relax}
\providecommand{\BIBentryALTinterwordstretchfactor}{4}
\providecommand{\BIBentryALTinterwordspacing}{\spaceskip=\fontdimen2\font plus
\BIBentryALTinterwordstretchfactor\fontdimen3\font minus
  \fontdimen4\font\relax}
\providecommand{\BIBforeignlanguage}[2]{{%
\expandafter\ifx\csname l@#1\endcsname\relax
\typeout{** WARNING: IEEEtran.bst: No hyphenation pattern has been}%
\typeout{** loaded for the language `#1'. Using the pattern for}%
\typeout{** the default language instead.}%
\else
\language=\csname l@#1\endcsname
\fi
#2}}
\providecommand{\BIBdecl}{\relax}
\BIBdecl

\bibitem{Huang-SPL-Soft-IoU-2019}
Y.~Huang, Z.~Tang, D.~Chen, K.~Su, and C.~Chen, ``Batching soft iou for
  training semantic segmentation networks,'' \emph{IEEE Signal Process. Lett.},
  vol.~27, pp. 66--70, 2019.

\bibitem{Li-SPL-Proposal-2019}
J.~Li, S.~He, H.-C. Wong, and S.-L. Lo, ``Proposal-driven segmentation for
  videos,'' \emph{IEEE Signal Process. Lett.}, vol.~26, no.~8, pp. 1098--1102,
  2019.

\bibitem{Arnab-SPM-CRF-2018}
A.~Arnab, S.~Zheng, S.~Jayasumana, B.~Romera-Paredes, M.~Larsson, A.~Kirillov,
  B.~Savchynskyy, C.~Rother, F.~Kahl, and P.~H. Torr, ``Conditional random
  fields meet deep neural networks for semantic segmentation: Combining
  probabilistic graphical models with deep learning for structured
  prediction,'' \emph{IEEE Signal Process. Mag.}, vol.~35, no.~1, pp. 37--52,
  2018.

\bibitem{Zhou-WWW-Multi-scale-2019}
Q.~Zhou, W.~Yang, G.~Gao, W.~Ou, H.~Lu, J.~Chen, and L.~J. Latecki,
  ``Multi-scale deep context convolutional neural networks for semantic
  segmentation,'' \emph{World Wide Web}, vol.~22, no.~2, pp. 555--570, 2019.

\bibitem{Wang-CVPR-DSR-2020}
L.~Wang, D.~Li, Y.~Zhu, L.~Tian, and Y.~Shan, ``Dual super-resolution learning
  for semantic segmentation,'' in \emph{Proc. IEEE Conf. Comput. Vis. Pattern
  Recognit.}, 2020, pp. 3774--3783.

\bibitem{Badrinarayanan-PAMI-SegNet-2017}
V.~Badrinarayanan, A.~Kendall, and R.~Cipolla, ``{SegNet}: A deep convolutional
  encoder-decoder architecture for image segmentation,'' \emph{IEEE Trans.
  Pattern Anal. Mach. Intell.}, vol.~39, no.~12, pp. 2481--2495, 2017.

\bibitem{Kitti}
A.~Geiger, P.~Lenz, and R.~Urtasun, ``Are we ready for autonomous driving? the
  kitti vision benchmark suite,'' in \emph{Proc. IEEE Conf. Comput. Vis.
  Pattern Recognit.}, 2012, pp. 3354--3361.

\bibitem{Cityscapes}
M.~Cordts, M.~Omran, S.~Ramos, T.~Rehfeld, M.~Enzweiler, R.~Benenson,
  U.~Franke, S.~Roth, and B.~Schiele, ``The cityscapes dataset for semantic
  urban scene understanding,'' in \emph{Proc. IEEE Conf. Comput. Vis. Pattern
  Recognit.}, 2016.

\bibitem{Zhou-ASC-AGLNet-2020}
Q.~Zhou, Y.~Wang, Y.~Fan, X.~Wu, S.~Zhang, B.~Kang, and L.~J. Latecki,
  ``Aglnet: Towards real-time semantic segmentation of self-driving images via
  attention-guided lightweight network,'' \emph{Applied Soft Computing},
  vol.~96, p. 106682, 2020.

\bibitem{Liang-PAMI-lip-2018}
X.~Liang, K.~Gong, X.~Shen, and L.~Lin, ``Look into person: Joint body parsing
  \& pose estimation network and a new benchmark,'' \emph{IEEE Trans. Pattern
  Anal. Mach. Intell.}, vol.~41, no.~4, pp. 871--885, 2018.

\bibitem{Krizhevsky-NIPS-Imagenet}
A.~Krizhevsky, I.~Sutskever, and G.~E. Hinton, ``Imagenet classification with
  deep convolutional neural networks,'' in \emph{Proc. Adv. Neural Inf.
  Process. Syst.}, 2012.

\bibitem{Simonyan-ICLR-VGG-2015}
K.~Simonyan and A.~Zisserman, ``Very deep convolutional networks for
  large-scale image recognition,'' \emph{Proc. Int. Conf. on Learn.
  Represent.}, 2015.

\bibitem{He-CVPR-ResNet-2016}
K.~He, X.~Zhang, S.~Ren, and J.~Sun, ``Deep residual learning for image
  recognition,'' in \emph{Proc. IEEE Conf. Comput. Vis. Pattern Recognit.},
  2016.

\bibitem{Huang-CVPR-DenseNet-2017}
G.~Huang, Z.~Liu, L.~van~der Maaten, and K.~Q. Weinberger, ``Densely connected
  convolutional networks,'' \emph{Proc. IEEE Conf. Comput. Vis. Pattern
  Recognit.}, pp. 2261--2269, 2017.

\bibitem{Long-CVPR-FCN-2015}
J.~Long, E.~Shelhamer, and T.~Darrell, ``Fully convolutional networks for
  semantic segmentation,'' in \emph{Proc. IEEE Conf. Comput. Vis. Pattern
  Recognit.}, 2015.

\bibitem{Chen-ICLR-Deeplab-2016}
L.-C. Chen, G.~Papandreou, I.~Kokkinos, K.~Murphy, and A.~L. Yuille, ``Semantic
  image segmentation with deep convolutional nets and fully connected crfs,''
  \emph{Proc. Int. Conf. on Learn. Represent.}, 2015.

\bibitem{Dai-ICCV-DCN-2017}
J.~Dai, H.~Qi, Y.~Xiong, Y.~Li, G.~Zhang, H.~Hu, and Y.~Wei, ``Deformable
  convolutional networks,'' in \emph{Proc. IEEE Int. Conf. Comput. Vis.}, 2017,
  pp. 764--773.

\bibitem{Yu-ECCV-RepGraph-2020}
C.~Yu, Y.~Liu, C.~Gao, C.~Shen, and N.~Sang, ``Representative graph neural
  network,'' in \emph{Proc. Eur. Conf. on Comput. Vis.}\hskip 1em plus 0.5em
  minus 0.4em\relax Springer, 2020, pp. 379--396.

\bibitem{Zhao-CVPR-PSPNet-2017}
H.~Zhao, J.~Shi, X.~Qi, X.~Wang, and J.~Jia, ``Pyramid scene parsing network,''
  \emph{Proc. IEEE Conf. Comput. Vis. Pattern Recognit.}, 2017.

\bibitem{Chen-Arxiv-Deeplabv3-2017}
L.-C. Chen, G.~Papandreou, F.~Schroff, and H.~Adam, ``Rethinking atrous
  convolution for semantic image segmentation,'' \emph{arXiv}, 2017.

\bibitem{Fu-TNNLS-DANet-2020}
J.~{Fu}, J.~{Liu}, J.~{Jiang}, Y.~{Li}, Y.~{Bao}, and H.~{Lu}, ``Scene
  segmentation with dual relation-aware attention network,'' \emph{IEEE Trans.
  Neural Netw. Learn. Syst.}, pp. 1--14, 2020.

\bibitem{Wang-CVPR-Nonlocal-2018}
X.~Wang, R.~Girshick, A.~Gupta, and K.~He, ``Non-local neural networks,''
  \emph{Proc. IEEE Conf. Comput. Vis. Pattern Recognit.}, 2018.

\bibitem{Yu-CVPR-CPNet-2020}
C.~Yu, J.~Wang, C.~Gao, G.~Yu, C.~Shen, and N.~Sang, ``Context prior for scene
  segmentation,'' in \emph{Proc. IEEE Conf. Comput. Vis. Pattern Recognit.},
  2020, pp. 12\,416--12\,425.

\bibitem{Yu-CVPR-DFN-2018}
C.~Yu, J.~Wang, C.~Peng, C.~Gao, G.~Yu, and N.~Sang, ``Learning a
  discriminative feature network for semantic segmentation,'' in \emph{Proc.
  IEEE Conf. Comput. Vis. Pattern Recognit.}, 2018.

\bibitem{Hu-CVPR-SEnet-2017}
J.~Hu, L.~Shen, and G.~Sun, ``Squeeze-and-excitation networks,'' \emph{Proc.
  IEEE Conf. Comput. Vis. Pattern Recognit.}, 2018.

\bibitem{Yu-ECCV-BiSeNet-2018}
C.~Yu, J.~Wang, C.~Peng, C.~Gao, G.~Yu, and N.~Sang, ``Bisenet: Bilateral
  segmentation network for real-time semantic segmentation,'' in \emph{Proc.
  Eur. Conf. on Comput. Vis.}, 2018, pp. 325--341.

\bibitem{Yu-ARXIV-BiSeNetV2-2020}
C.~Yu, C.~Gao, J.~Wang, G.~Yu, C.~Shen, and N.~Sang, ``Bisenet v2: Bilateral
  network with guided aggregation for real-time semantic segmentation,''
  \emph{arXiv}, 2020.

\bibitem{Xu-NIPS-DFN-2016}
X.~Jia, B.~De~Brabandere, T.~Tuytelaars, and L.~V. Gool, ``Dynamic filter
  networks,'' in \emph{Proc. Adv. Neural Inf. Process. Syst.}, 2016, pp.
  667--675.

\bibitem{Su-CVPR-PAC-2019}
H.~Su, V.~Jampani, D.~Sun, O.~Gallo, E.~Learned-Miller, and J.~Kautz,
  ``Pixel-adaptive convolutional neural networks,'' in \emph{Proc. IEEE Conf.
  Comput. Vis. Pattern Recognit.}, 2019, pp. 11\,166--11\,175.

\bibitem{Tian-ECCV-CondINS-2020}
Z.~Tian, C.~Shen, and H.~Chen, ``Conditional convolutions for instance
  segmentation,'' in \emph{Proc. Eur. Conf. on Comput. Vis.}, 2020.

\bibitem{Ding-CVPR-SVCNet-2019}
H.~Ding, X.~Jiang, B.~Shuai, A.~Q. Liu, and G.~Wang, ``Semantic correlation
  promoted shape-variant context for segmentation,'' in \emph{Proc. IEEE Conf.
  Comput. Vis. Pattern Recognit.}, 2019, pp. 8885--8894.

\bibitem{Yang-NIPS-CondConv-2019}
B.~Yang, G.~Bender, Q.~V. Le, and J.~Ngiam, ``Condconv: Conditionally
  parameterized convolutions for efficient inference,'' in \emph{Proc. Adv.
  Neural Inf. Process. Syst.}, 2019, pp. 1307--1318.

\bibitem{Chen-CVPR-DynamicConv-2020}
Y.~Chen, X.~Dai, M.~Liu, D.~Chen, L.~Yuan, and Z.~Liu, ``Dynamic convolution:
  Attention over convolution kernels,'' in \emph{Proc. IEEE Conf. Comput. Vis.
  Pattern Recognit.}, 2020, pp. 11\,030--11\,039.

\bibitem{Milletari-3DV-VNet-2016}
F.~Milletari, N.~Navab, and S.-A. Ahmadi, ``V-net: Fully convolutional neural
  networks for volumetric medical image segmentation,'' in \emph{Int. Conf. on
  3D Vis.}\hskip 1em plus 0.5em minus 0.4em\relax IEEE, 2016, pp. 565--571.

\bibitem{Zhou-ADE-2016}
B.~Zhou, H.~Zhao, X.~Puig, S.~Fidler, A.~Barriuso, and A.~Torralba, ``Semantic
  understanding of scenes through the ade20k dataset,'' \emph{Int. J. Comput.
  Vis.}, vol. 127, pp. 302--321, 2018.

\bibitem{PASCAL-Context}
R.~Mottaghi, X.~Chen, X.~Liu, N.-G. Cho, S.-W. Lee, S.~Fidler, R.~Urtasun, and
  A.~Yuille, ``The role of context for object detection and semantic
  segmentation in the wild,'' in \emph{Proc. IEEE Conf. Comput. Vis. Pattern
  Recognit.}, 2014.

\bibitem{Chen-ECCV-Deeplabv3p-2018}
L.-C. Chen, Y.~Zhu, G.~Papandreou, F.~Schroff, and H.~Adam, ``Encoder-decoder
  with atrous separable convolution for semantic image segmentation,'' in
  \emph{Proc. Eur. Conf. on Comput. Vis.}, 2018, pp. 801--818.

\bibitem{Xiao-UperNet-ECCV-2018}
T.~Xiao, Y.~Liu, B.~Zhou, Y.~Jiang, and J.~Sun, ``Unified perceptual parsing
  for scene understanding,'' in \emph{Proc. Eur. Conf. on Comput. Vis.}, 2018,
  pp. 418--434.

\bibitem{Zhao-ECCV-PSANet-2018}
H.~Zhao, Y.~Zhang, S.~Liu, J.~Shi, C.~C. Loy, D.~Lin, and J.~Jia, ``{PSANet}:
  Point-wise spatial attention network for scene parsing,'' in \emph{Proc. Eur.
  Conf. on Comput. Vis.}, 2018.

\bibitem{Zhang-CVPR-EncNet-2018}
H.~Zhang, K.~Dana, J.~Shi, Z.~Zhang, X.~Wang, A.~Tyagi, and A.~Agrawal,
  ``Context encoding for semantic segmentation,'' in \emph{Proc. IEEE Conf.
  Comput. Vis. Pattern Recognit.}, 2018, pp. 7151--7160.

\bibitem{Zhang-CVPR-CFNet-2019}
H.~Zhang, H.~Zhang, C.~Wang, and J.~Xie, ``Co-occurrent features in semantic
  segmentation,'' in \emph{Proc. IEEE Conf. Comput. Vis. Pattern Recognit.},
  2019, pp. 548--557.

\bibitem{Zhu-ICCV-ANL-2019}
Z.~Zhu, M.~Xu, S.~Bai, T.~Huang, and X.~Bai, ``Asymmetric non-local neural
  networks for semantic segmentation,'' in \emph{Proc. IEEE Int. Conf. Comput.
  Vis.}, 2019, pp. 593--602.

\bibitem{Yuan-ECCV-OCRNet-2019}
Y.~Yuan, X.~Chen, and J.~Wang, ``Object-contextual representations for semantic
  segmentation,'' in \emph{Proc. Eur. Conf. on Comput. Vis.}, 2020.

\bibitem{He-CVPR-APCNet-2019}
J.~He, Z.~Deng, L.~Zhou, Y.~Wang, and Y.~Qiao, ``Adaptive pyramid context
  network for semantic segmentation,'' in \emph{Proc. IEEE Conf. Comput. Vis.
  Pattern Recognit.}, 2019, pp. 7519--7528.

\bibitem{Zhang-ARXIV-ResNest-2020}
H.~Zhang, C.~Wu, Z.~Zhang, Y.~Zhu, Z.~Zhang, H.~Lin, Y.~Sun, T.~He, J.~Mueller,
  R.~Manmatha \emph{et~al.}, ``Resnest: Split-attention networks,'' \emph{arXiv
  preprint arXiv:2004.08955}, 2020.

\bibitem{Ding-CVPR-CCL-2018}
H.~Ding, X.~Jiang, B.~Shuai, A.~Qun~Liu, and G.~Wang, ``Context contrasted
  feature and gated multi-scale aggregation for scene segmentation,'' in
  \emph{Proc. IEEE Conf. Comput. Vis. Pattern Recognit.}, 2018, pp. 2393--2402.

\bibitem{Fu-CVPR-DANet-2019}
J.~Fu, J.~Liu, H.~Tian, Z.~Fang, and H.~Lu, ``Dual attention network for scene
  segmentation,'' \emph{Proc. IEEE Conf. Comput. Vis. Pattern Recognit.}, 2019.

\end{thebibliography}
\end{document}